\UseRawInputEncoding

\documentclass[journal]{IEEEtran}
\usepackage{cite}
\usepackage{amsmath,amssymb}

\usepackage{multirow}
\usepackage{algorithm}
\usepackage{algorithmic}
\usepackage{graphicx}
\usepackage{bm}
\usepackage[T1]{fontenc}
\usepackage{diagbox}
\ifCLASSINFOpdf
\else
\fi
\begin{document}
%
\title{Unified Feature Extraction Framework based on Contrastive Learning}
%
%
%

\author{Hongjie~Zhang, Wenwen~Qiang, 
        Jinxin~Zhang,
        and~Ling~Jing
\thanks{This work was supported by Next Generation Precision Aquaculture: R\&D on intelligent measurement, control technology (No. 2017YFE0122100), the National Natural Science Foundation of China (Nos. 62076244, 12071024).}%

\thanks{H. Zhang is associated with the College of Information and Electrical Engineering, China Agricultural University, Beijing 100083, PR China; Beijing Engineering and Technology Research Center for Internet of Things in Agriculture, Beijing 100083, PR China. }

\thanks{W. Qiang is associated with the University of Chinese Academy of Sciences, Beijing, PR China, in addition to the Science \& Technology on Integrated Information System Laboratory, Institute of Software Chinese Academy of Sciences, Beijing, PR China.}


\thanks{J. Zhang is associated with the College of Information and Electrical Engineering, China Agricultural University, Beijing 100083, PR China; National Innovation Center for Digital Fishery, China Agricultural University, Beijing 100083, PR China. }

\thanks{L. Jing is associated with the College of Science, China Agricultural University, Beijing 100083, PR China; Key Laboratory of Agricultural Information Acquisition Technology, Ministry of Agriculture, Beijing 100083, PR China. (Corresponding author: Ling Jing, email: jingling@cau.edu.cn)}}
\maketitle

\begin{abstract}
Feature extraction is an efficient approach for alleviating the issue of dimensionality in high-dimensional data. As a popular self-supervised learning method, contrastive learning has recently garnered considerable attention. In this study, we proposed a unified framework based on a new perspective of contrastive learning (CL) that is suitable for both unsupervised and supervised feature extraction. The proposed framework first constructed two CL graph for uniquely defining the positive and negative pairs. Subsequently, the projection matrix was determined by minimizing the contrastive loss function. In addition, the proposed framework considered both similar and dissimilar samples to unify unsupervised and supervised feature extraction. Moreover, we propose the three specific methods: unsupervised contrastive learning method, supervised contrastive learning method 1 ,and supervised contrastive learning method 2. Finally, the numerical experiments on five real datasets demonstrated the superior performance of the proposed framework in comparison to the existing methods.
\end{abstract}


%
\IEEEpeerreviewmaketitle

\section{Introduction}
%
%
%
%
\IEEEPARstart{C}{urrently}, high-dimensional data is widely used in pattern recognition and data mining\cite{1,2,3}, which wastes considerable time and cost apart from causing the problem known as ``curse of dimensionality''\cite{4}. Thus, feature extraction of data poses immense significance to solve this issue\cite{ 5,6,40,7}.\par

Recently, self-supervised learning\cite{29,30,31} has become a popular topic in the field of deep learning. Self-supervised learning is a method of unsupervised learning that uses data information to supervise itself by constructing positive and negative pairs. Self-supervised learning strives to learn more discriminative features to effectively bridge unsupervised and supervised learning. Consequently, contrastive learning\cite{32,33,34,35} has attracted extensive scholarly attention as the primary method of self-supervised learning. Tian et al. proposed contrastive multiview coding (CMC) to process multiview data\cite{36}. First, CMC constructs the same samples in any two views as positive pairs and dissimilar samples as negative pairs, and subsequently optimizes a neural network framework by minimizing the contrastive loss function to maximize the similarity of the projected positive pairs. In addition, Chen et al. proposed a simple framework for contrastive learning (SimCLR) \cite{37} to process the problem. First, it performs data enhancement to obtain distinct representations of the same sample to consider them as positive pairs, and thereafter, considers the representations of any two distinct samples as negative pairs. Finally, SimCLR optimizes the network framework by minimizing the contrastive loss, similar to CMC. However, for the problem of single view feature extraction, the existing models based on contrastive learning still have many defects. For example, data enhancement will increase the running time of the algorithm, and defining all different samples as negative pairs will lead to the separation of samples of the same class.\par

 Inspired by our prior research, we proposed a unified feature extraction framework based on contrastive learning (CL-UFEF) that is suitable for both unsupervised and supervised feature extraction for single-view data. The proposed framework, CL-UFEF, constructed two contrastive graphs (CLG) to establish a new approach for defining positive and negative pairs, instead of conducting data enhancement. Moreover, the contrastive loss function considers both similar and dissimilar samples based on the CLG to unify the aspects of unsupervised and supervised feature extraction. Furthermore, the effectiveness of the proposed framework was verified by three specific models, including the unsupervised contrastive learning method (u-CL), supervised contrastive learning method 1 (s-CL1), and supervised contrastive learning method 2 (s-CL2), respectively. \par
The main contributions of this study are as follows:
\begin{itemize}
	\item A unified feature extraction framework for single-view data based on contrastive learning (CL-UFEF) is proposed from a new perspective that is suitable for both unsupervised and supervised cases.
	\item CL-UFEF proposes a novel approach to define positive and negative pairs in contrastive learning.
	\item Based on CL-UFEF, we propose three specific models, including u-CL, s-CL1, and s-CL2, and the experiments on five real datasets proved the advantages of the proposed framework.
\end{itemize}
\par 

The remainder of this article is organized as follows. The traditional feature extraction methods are briefly introduced in Section \uppercase\expandafter{\romannumeral2}. Subsequently, the development of the unified feature extraction framework (CL-UFEF) and the u-CL, s-CL1, and s-CL2 models are discussed in Section \uppercase\expandafter{\romannumeral3}. In addition, the extensive experiments conducted on several real-world datasets are presented in Section \uppercase\expandafter{\romannumeral4}. Finally, the conclusions of the current study are detailed in Section \uppercase\expandafter{\romannumeral5}.

\section{Related Work}

Recently, several feature extraction methods\cite{8,9,10,11,12} have been proposed, which can be categorized into unsupervised\cite{13}, semi-supervised\cite{14}, and supervised\cite{15} methods. In this study, we focused on unsupervised and supervised feature extraction. In the unsupervised feature extraction models, classical principal-component analysis (PCA)\cite{8} provides a wide range of applicability and effectiveness, as it seeks the maximum variance of samples in the subspace, which is more conducive to subsequent classification, clustering, or other tasks. Nonetheless, PCA is a linear feature extraction method that is unsuitable for handling nonlinear data. With the development of manifold learning, numerous nonlinear feature extraction methods such as isometric feature mapping (ISOMAP)\cite{16}, Laplacian eigenmap (LE)\cite{17}, and local linear embedding (LLE)\cite{18} have been proposed to solve this problem. However, these nonlinear feature extraction methods cannot be applied to new sample points because they directly obtain the low-dimensional representation of samples without using a projection matrix. Therefore, several nonlinear methods based on manifold assumptions have been modified into linearized versions. 
In particular, locality preserving projections (LPP)\cite{19}, neighborhood preserving embedding (NPE)\cite{20}, and isometric projection are considered as linearized LE, LLE, and ISOMAP, respectively. Although all the feature extraction methods preserve the manifold structures in subspace, they have various requirements for manifold learning. For instance, LPP obtains the neighbor graph of the original data in advance, and then the samples in the subspace maintain the same neighbor relationship. In contrast, NPE expects that the linear reconstruction relationship of the original space can be maintained between the samples and neighborhood samples after feature extraction. However, the above graph-based methods only considered the local structures of the data and neglected the global structures. Thus, sparsity preserving projections (SPP)\cite{21}, collaborative representation-based projections (CRP)\cite{22}, and low-rank preserving embedding (LRPE)\cite{23} have been proposed to effectively solve this problem. SPP constructs a $l1$-graph with adaptive neighbors by utilizing the sparsity technique, whereas CRP aims to build a $l2$-graph by calculating the linear reconstruction coefficients of each sample based on the remaining samples; LRPE constructs a nuclear norm graph with adaptive neighbors using low-rank representation. However, the unsupervised graph-based methods closely projects the original similar samples in the subspace and neglects the dissimilar samples.\par 

Supervised feature extraction methods obtain more discriminant information using sample labels. For instance, linear discriminant analysis (LDA)\cite{25}  seeks an embedding transformation such that the within-class scatter is minimized and the between-class scatter is maximized. However, LDA, similar to PCA, is a linear feature extraction method as well. Consequently, it might deliver poor performance if the samples in a class form several separate clusters (i.e., multi-mode). Thus, researchers have proposed using local Fisher discriminant analysis (LFDA)\cite{26} and marginal Fisher analysis (MFA)\cite{24} based on GE and utilizing the approach of manifold learning in unsupervised feature extraction models. LFDA combines the ideas of LDA and LPP to locally construct the levels of within- and between-class scatters, which allows the LFDA to simultaneously achieve maximum preservation of the within- and between-class local structures at the same time. MFA is distinct from the LFDA as it considers the local structure within classes and constructs the local structure relationship between classes by accounting for the samples on the edges of various classes. However, MFA is limited by the problem of class isolation, i.e., not all samples of heterogeneous edges have local neighbor relationships. In this context, researchers have proposed multiple marginal Fisher analysis (MMFA)\cite{27}, which selects the nearest neighbor samples on all heterogeneous edges to construct the local relationship between classes. Sparsity preserving discriminant projections (SPDP)\cite{28} is proposed based on SPP to maintain the sparse reconstruction coefficients of samples in the subspace. Note that these graph-based methods initially learn an affine graph using various measure metrics, and subsequently evaluate the projection based on graphs.\par 

Inspired by traditional graph-based methods, we proposed a unified feature extraction framework based on contrastive learning (CL-UFEF). Concretelly, compared with the previous models based on contrastive learning, CL-UFEF does not need data enhancement, and it constructs positive and negative pairs based on two contrastive learning graphs (CLG), which will make the similar samples in the subspace more clustered. Compared with the traditional graph-based models, CL-UFEF is suitable for both unsupervised and supervised feature extraction, and it considers similar and dissimilar samples in unsupervised and spervised learning based on the CLG.

\section{Methodology}
\begin{figure*}[!ht]
	\centering
	\includegraphics[width=1\textwidth]{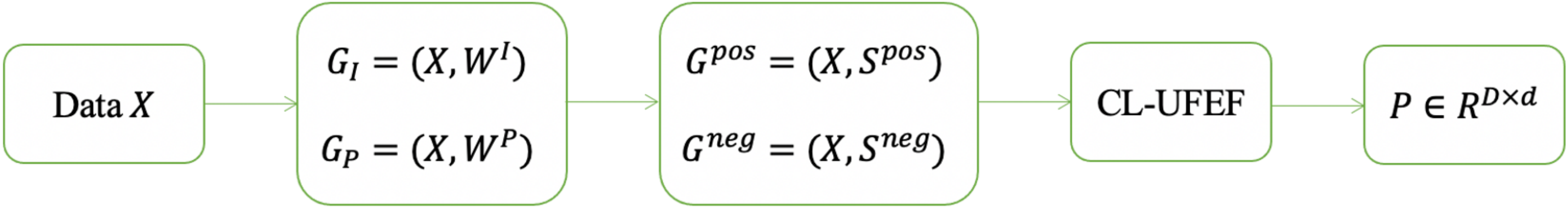}
	\caption{Process of determining projection matrix $P$ using CL-UFEF.} 
	\label{fig}
\end{figure*}

\begin{table}[!ht]      
	\centering
	\caption{Notations and definitions.}
	\label{Table1}
	\begin{tabular}[t]{l l}
		\hline
		$X$ & Training sample set\\
		$Y$ & Set of training samples in a low-dimensional space\\
		$n$ & Number of training samples\\
		$D$ & Dimensionality of the samples in the original space\\
		$d$ & Dimensionality of embedding features\\
		$c_i$ & Labels of samples $x_i$\\
		$C$ & Number of classes\\
		$P$ & Projection matrix\\
		$CLG$& Contrastive learning graph\\
		$G^{pos}$ & Positive graph\\
		$G^{neg}$ & Negative graph\\
		$S^{pos}$ & Similarity matrix for positive pairs\\
		$S^{neg}$ & Dissimilarity matrix of negative pairs\\
		$\sigma$ & Positive parameter\\
		$k$ & Number of neighbors\\
		$NK(x_j)$ &The $k$ nearest neighbors of $x_j$\\
		$NK^+_{(x_j)}$ &The $k$ nearest neighbors of $x_j$ in the $c_j$ th class\\
		$\nabla L(P)$ & Gradient of $L(P)$ with respect to $P$\\
		$T$ & Number of iterations\\
		\hline
	\end{tabular}
\end{table}

In this section, a unified feature extraction framework is proposed based on contrastive learning (CL-UFEF) for both unsupervised and supervised feature extraction, including three specific cases: u-CL, s-CL1, and s-CL2. The projection matrix evaluation process using CL-UFEF is intuitively illustrated in Figure \ref{fig}.\par

This study considered the problem of unsupervised and supervised feature extraction. Let us mathematically formulate the feature extraction problem as follows. \par  
Feature extraction problem:  Given a training sample set $X=[x_1,x_2, . . .,x_n]\in{R^{D\times n}}$, where $n$ and $D$ are the number of samples and features, respectively. In the supervised case, the class label $c_i$ of $x_i$ was provided, where $c_i\in\{1,2,...,C\}$ and $C$ represent the number of classes. The purpose of feature extraction is to find a projection matrix $P\in{R^{D\times d}}$ to derive the low-dimensional embedding $Y = [y_1, y_2, . . ., y_n]\in{R^{d\times n}}$ for $X$ calculated by $Y = P^TX$, where $d\ll D$.\par 
For convenience, the symbols used in this study are summarized in Table \ref{Table1}.\par

\subsection{Unified Feature Extraction Framework based on Contrastive Learning (CL-UFEF)} 

\subsection*{Construct CLG}
The contrastive learning was applied by constructing the two contrastive learning graphs (CLG): a positive graph $G ^ {pos} = \{X, {S^{pos}} \}$ and a negative graph $G ^ {neg} = \{X, {S^{neg}} \} $, which  define the positive and negative pairs, including the positive matrix $S^{pos}=(S^{pos}_{i,j})_{n\times n}$ and the negative matrix $S^{neg}=(S^{neg}_{i,j})_{n\times n}$ to measure the similarity of the positive pairs and the dissimilarity of the negative pairs. Morever, it should be note that the greater number of negative pairs strengthens the influence of the model based on contrastive learning\cite{38}.
\par 

In particular, compared with the previous methods based on ciontrastive learning, $G^{pos}$ and $G^{neg}$ define similar samples as positive pairs and dissimilar samples as negative pairs. Compared with the traditional graph-based methods, $G^{pos}$ and $G^{neg}$ are suitable for both supervised and unsupervised learning, and dissimilar samples are considered in unsupervised and spervised learning based on the $G^{neg}$.

\subsection*{Model of CL-UFEF}
Contrastive learning is realized using the positive matrix $S^{pos}$ and negative matrix $S^{neg}$. We hope that the projections of the positive pair $x_i$ and $x_j$ with larger $S^{pos}_{i,j}$ should have greater similarity, and the projections of the negative pair $x_i$ and $x_j$ with larger $S^{neg}_{i,j}$ will have greater dissimilarity. Thus, the optimization problem of the CL-UFEF is proposed as follows: 
\begin{center}
	\begin{equation}\label{CL-UFEF}
	\min_{P}L(P)=\sum_{i=1}^{n}-log\frac{ \sum_{j=1}^{n}S^{pos}_{i,j}exp(SIM(P^Tx_i,P^Tx_j))}{\sum_{j=1}^{n}S^{who}_{i,j}exp(SIM(P^Tx_i,P^Tx_j))}
	\end{equation}
\end{center}
where，
\begin{center}
	\begin{equation}
	SIM(P^Tx_i,P^T,x_j)=\frac{x_i^TPP^Tx_j}{\|P^Tx_i\|\|P^Tx_j\|\sigma}
	\end{equation}
\end{center}
$\sigma$ is a positive parameter, and $S^{who}_{i,j}=S^{pos}_{i,j}+S^{neg}_{i,j}$ is the whole similarity matrix. \par 
 
Subsequently, based on the specific construction methods of  $G^{pos}$ and $G^{neg}$, three special cases, including u-CL, s-CL1, and s-CL2 are proposed to verify the effectiveness of the proposed framework. Morever, some other construction methods of $G^{pos}$ and $G^{neg}$ are presented in APPENDIX A.

\subsection{Unspervised contrastive learning method (u-CL)}
We use the k-nearest neighbor method to construct CLG, indicating that the local $k$ nearest neighbors were positive pairs whereas the non-nearest neighbors were negative pairs. 
In particular, the positive matrix $S^{pos}$ and negative matrix $S^{neg}$ can be defined as follows:
\begin{center}
	\begin{equation}\label{que1}
	S^{pos}_{i,j}=\left\{
	\begin{aligned}
	&exp(\frac{-||x_i-x_j||_2^2}{t}), &{\rm if} \, x_i\in NK(x_j)\\
	& &{\rm or}  \, x_j\in NK(x_i);\\
	&0, &{\rm otherwise},
	\end{aligned}
	\right.
	\end{equation}
\end{center}

\begin{center}
	\begin{equation}\label{que2}
	S^{neg}_{i,j}=\left\{
	\begin{aligned}
	&1, &{\rm if} \, x_i\notin NK(x_j)\,{\rm and}\, x_j\notin NK(x_i); \\
	&0, &{\rm otherwise}.
	\end{aligned}
	\right.
	\end{equation}
\end{center}
where $t$ is the thermal parameter used for adjusting the value range of the weight matrix $S^{pos}$, $NK(x_j)$ represents the $k$ nearest neighbors of $x_j$ , and the parameter $k$ is a tuning parameter. The optimization problem of u-CL can be obtained by introducing (\ref{que1}) and (\ref{que2}) into CL-UFEF.
\subsection{Spervised contrastive learning method 1 (s-CL1)}
We use class information samples to construct CLG, indicating that the within-class samples are positive pairs and between-class samples are negative pairs. In particular, the positive matrix $S^{pos}$ and negative matrix $S^{neg}$ can be defined as 
\begin{center}
	\begin{equation}\label{que3}
	S^{pos}_{i,j}=\left\{
	\begin{aligned}
	&1, &{\rm if}\, c_i=c_j;\\
	&0, &{\rm if}\, c_i\neq c_j,
	\end{aligned}
	\right.
	\end{equation}
\end{center}

\begin{center}
	\begin{equation}\label{que4}
	S^{neg}_{i,j}=\left\{
	\begin{aligned}
	&1, &{\rm if}\,  c_i\neq c_j;\\
	&0, &{\rm if}\,  c_i=c_j.
	\end{aligned}
	\right.
	\end{equation}
\end{center}
Moreover, the optimization problem of s-CL1 can be obtained by introducing (\ref{que3}) and (\ref{que4}) into the CL-UFEF.

\subsection{Spervised contrastive learning method 2 (s-CL2)}
We use class information and k nearest neighbors of samples to construct CLG, indicating that local $k$ nearest neighbors in the same class were positive pairs and the samples of other relationships were negative pairs. In particular, the positive matrix $S^{pos}$ and negative matrix $S^{neg}$ can be defined as
\begin{center}
	\begin{equation}\label{que5}
	S^{pos}_{i,j}=\left\{
	\begin{aligned} 
	&exp(\frac{-||x_i-x_j||_2^2}{t}), &{\rm if} \,x_i\in NK^+(x_j)\\
	&& {\rm or} \, x_j\in NK^+(x_i);\\
	&0, &{\rm otherwise},
	\end{aligned}
	\right.
	\end{equation}
\end{center}

\begin{center}
	\begin{equation}\label{que6}
	S^{neg}_{i,j}=\left\{
	\begin{aligned}
	&1, &{\rm if} \, x_i\notin NK^+(x_j) \, {\rm and}\, x_j\notin NK^+(x_i);\\
	&0, &{\rm otherwise}.
	\end{aligned}
	\right.
	\end{equation}
\end{center}
where $NK^+_{(x_j)}$ represents the $k$ nearest neighbors of $x_j$ in the same class.
The optimization problem of s-CL2 can be obtained by introducing (\ref{que5}) and (\ref{que6}) into CL-UFEF.

\subsection{Optimization algorithm}

\begin{algorithm}[!h]\label{Algorithm1}
	\caption{CL-UFEF} 
	{\bf Input:} 
	
	Data matrix: $X\in R^{D\times n}$, $\alpha,\beta_1,\beta_2,\epsilon,P_0$.\\
	$m_0=0 $(Initialize $1^{st}$ moment vector)\\
	$v_0=0 $(Initialize $2^{nd}$ moment vector)\\
	$t= 0$ (Initialize time step)\\
	\hspace*{0.02in} {\bf Output:} 
	Projection matrix $P$
	\begin{algorithmic}
		\WHILE{$P_t$ not converged} 
		\STATE
		$t=t+1$\\
		$g_t=\bigtriangledown L(P_{t-1})$ is calculated using (\ref{grad}) (Obtain gradients with respect to the stochastic objective at time step $t$)\\
		$m_t=\beta_1\cdot m_{t-1}+(1-\beta_1)\cdot g_t$ (Update biased first-moment estimate)\\
		$v_t=\beta_2\cdot v_{t-1}+(1-\beta_2)\cdot g_t^2$ (Update biased second raw-moment estimate)\\
		$\hat{m}_t=m_t/(1-\beta_1^t)$ (Compute bias-corrected first-moment estimate)\\
		$\hat{v}_t=v_t/(1-\beta_2^t)$ (Compute bias-corrected second raw-moment estimate)\\
		$P_t=P_{t-1}-\alpha\cdot\hat{m}_t/(\sqrt{\hat{v}_t}+\epsilon)$\\
		\ENDWHILE
		\RETURN result
	\end{algorithmic}
\end{algorithm}

In this section, the optimization algorithm of CL-UFEF is presented (Algorithm 1), which uses the Adam optimizer\cite{39} to solve the optimization problem for CL-UFEF. Adam is an advancement on the random gradient descent method and can rapidly yield accurate results. This method calculates the adaptive learning rate of various parameters based on the budget of the first and second moments of the gradient. The parameters $\alpha $, $\beta_1$, $\beta_2$, and $\epsilon$ represent the learning rate, the exponential decay rate of the first- and second-order moment estimation, and the parameter to prevent division by zero in the implementation, respectively. In addition, the gradient of the loss function $L(P)$ with respect to the projection matrix $P$ was obtained from (\ref{grad}).\par 
The convergence condition of Algorithm 1 was set as $L(P_t)-L(P_{t-1})<0.001$, where $L(P_t)$ and $L(P_{t-1})$ are the function values obtained after the $t$th and $t-1$th gradient descent, respectively. Therefore, the computational complexity of Algorithm 1 is primarily executed in the first step, and the complexity of the derivative of the objective function is $O(Tn^3)$, where $T$ denotes the number of iterations.

\begin{center}
	\begin{equation}\label{grad} 
	\begin{aligned}
	&\nabla      L(P)=\\
	&\sum_{i=1}^{n} \{-\frac{\sum_{j=1}^{n} S^{who}_{i,j}exp(SIM(P^Tx_i,P^Tx_j))}{\sum_{j=1}^{n} S^{pos}_{i,j}exp(SIM(P^Tx_i,P^Tx  _j)}\cdot\\
	&[\sum_{j=1}^{n}S^{pos}_{i,j}exp(P^Tx_i,P^Tx_j)\nabla SIM(P^Tx_i,P^Tx_j)\cdot\\
	&\sum_{j=1}^{n}S^{who}_{i,j}exp(P^Tx_i,P^Tx_j)\\
	&-\sum_{j=1}^{n}S^{who}_{i,j}exp(P^Tx_i,P^Tx_j)\nabla SIM(P^Tx_i,P^Tx_j)\cdot\\
	&\sum_{j=1}^{n}S^{pos}_{i,j}exp(P^Tx_i,P^Tx_j)]/\\
	&[\sum_{j=1}^{n}S^{who}_{i,j}exp(P^Tx_i,P^Tx_j)]^2\},    
	\end{aligned}
	\end{equation}
\end{center}
where
\begin{center}
	\begin{equation}
	\begin{aligned}
	&\nabla SIM(P^Tx_i,P^Tx_j)=\\
	&\{(x_i{x_j}^T+x_j{x_i}^T)P\cdot\|P^Tx_i\|\|P^Tx_j\|\sigma\\
	&-[({x_i}^TPP^Tx_i)^{-\frac{1}{2}}\cdot x_i{x_i}^TP\cdot\|P^Tx_j\|\sigma\\
	&+({x_j}^TPP^Tx_j)^{-\frac{1}{2}}\cdot x_j{x_j}^TP\cdot\|P^Tx_i\|\sigma]\cdot {x_i}^TPP^Tx_j\}/\\
	&{(\|P^Tx_i\|\|P^Tx_j\|\sigma)}^2
	\end{aligned}
	\end{equation}
\end{center}

The main computational complexity of each cycle in Algorithm 1 is the derivation of the loss function in the first step, which is $O(n^2(D^2d+Dd+D^2))$. Assuming that the Algorithm 1 performs a total of $T$ cycles when converging, the main computational complexity is $O(n^2T(D^2d+Dd+D^2))$.

\section{Experimental results}

				\begin{figure*}[!ht]
					\centering
					\includegraphics[width=0.7\textwidth]{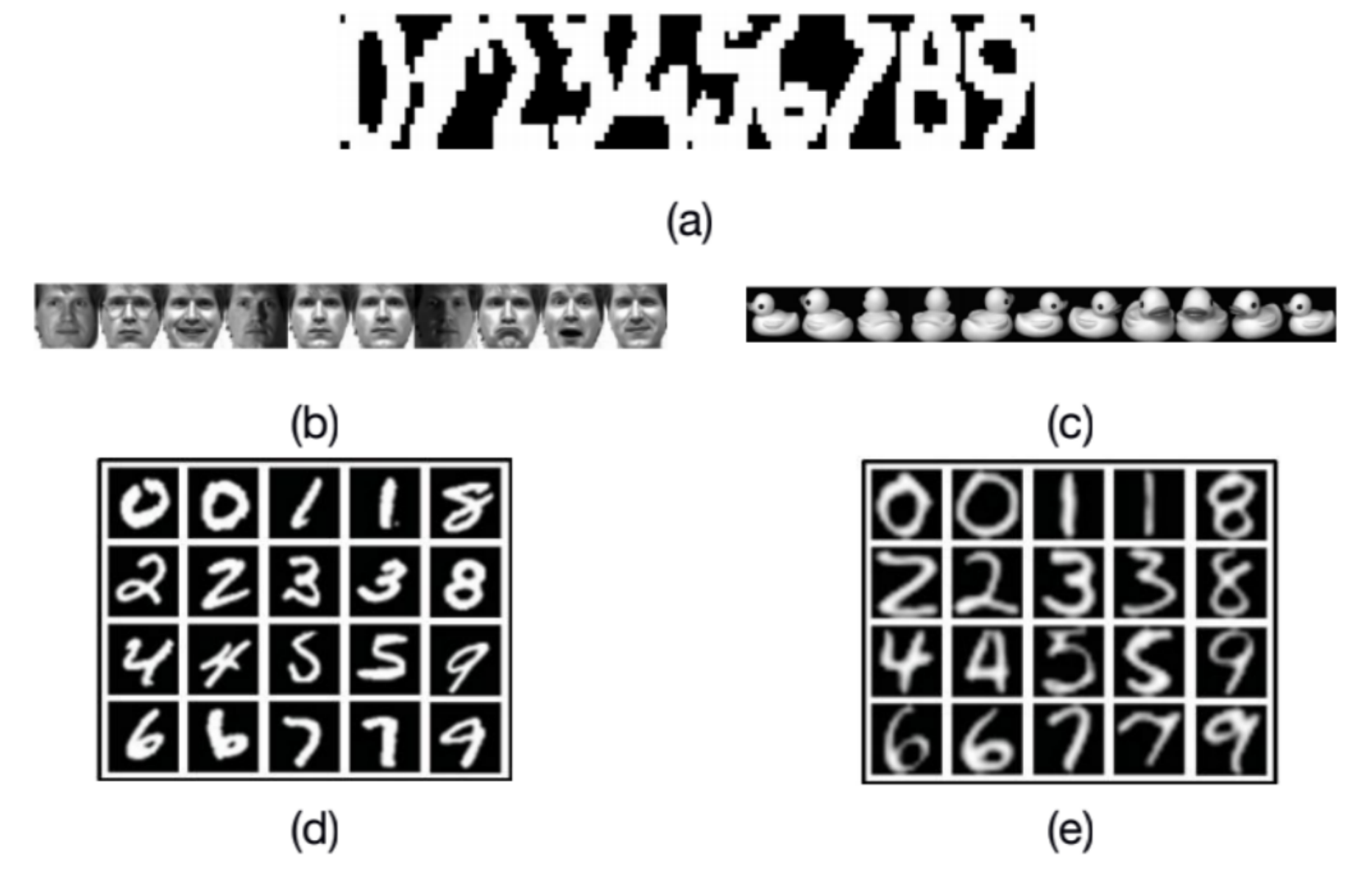}
					\caption{Image samples used in experiments: (a) Multiple features data set. (b)  Yale dataset. (c) COIL20 dataset. (d) MNIST data set. (e) USPS dataset.} 
					\label{fig0}
				\end{figure*}

\subsection{Data Descriptions and Experimental Setups}
In this study, the numerical experiments utilized five real datasets to demonstrate the performance advantages of the proposed CL-UFEF.\par 
Multiple Features dataset: This dataset was acquired from the UCI machine learning repository, comprising six distinct feature sets extracted from handwritten numbers from 0 to 9 with 200 patterns per class (i.e., 2000 patterns in ten classes). All the datasets were digitized in binary images. The six feature sets included fou (Fourier coefficients, 76 features), fac (profile correlations, 216 features), kar (Karhunen--Loeve coefficients, 64 features), pix (pixel averages in $2\times3$ windows, 240 features), zer (Zernike moments, 47 features), and mor (morphological features, six features). Example images are presented in Figure \ref{fig0} (a).\par 
Yale dataset: The dataset was created by Yale University Computer Vision and Control Center, containing data of $15$ individuals, wherein each person has 11 frontal images ($64\times64$ pixels in size) captured under various lighting conditions. The images were edited to $50\times40$ pixels with 256 Gy levels per pixel; examples are presented in Figure \ref{fig0} (b).\par 
COIL20 dataset: The COIL20 dataset was created by Columbia University in 1996, containing 1440 images of 20 objects, wherein each object includes 72 images ($64\times64$ pixels in size). In addition, each image was edited to $32\times32$ pixels, and each pixel had 256 Gy levels; examples are presented in Figure \ref{fig0} (c).\par 
MNIST dataset: This dataset contains 70,000 samples of $0--9$ digital images with a size of $28\times28$. We randomly selected 2000 images as experimental data, uniformly rescaled all the images to a size of $16\times16$, and used a feature vector of 256-level grayscale pixel values  to represent each image. Examples are presented in Figure \ref{fig0} (d).\par 
USPS dataset: This dataset contains 9298 samples of $0--9$ handwritten digital images, and the size of each image was adjusted to $16\times16$. A total of 1800 images were randomly selected as experimental data, and each image was represented by a feature vector of 256-level grayscale pixel values. Examples are depicted in Figure \ref{fig0} (e).\par 

In data processing, in order to shorten the running time, we first used PCA to reduce the dimensions of the Yale, COIL20, MNIST, and USPS datasets, and the details are listed in Table \ref{Table2}. Thereafter, we separately standardized these five datasets to improve the convergence speed of the model. Finally, we compared u-CL, s-CL1, and s-CL2 with the traditional graph-based methods of LPP, FLPP, LDA, LFDA, and FDLPP and the method SimCLR beased on contrastive learning to verify the advantages of our proposed feature extraction framework. Note that the $k$-nearest neighbor classifier ($k$ = 1) was used in the experiment. Moreover, six samples of each class were randomly selected from the Yale, COIL20, and each feature set of multiple feature datasets for training, and the remaining data were used for testing. Furthermore, nine samples of each class were randomly selected from MNIST and USPS datasets for training, and the remaining data were used for testing. All the processes were repeated five times, and the final evaluation criteria constituted the average recognition accuracy and average recall rate of five repeated experiments. The calculation method of recognition accuracy and recall rate are shown in (\ref{accuracy}) and (\ref{recall}). The experiments were implemented using MATLAB R2018a on a computer with an Intel Core i5-9400 2.90 GHz CPU and Windows 10 operating system.

\begin{equation}\label{accuracy}
	{\rm recognition\;accuracy}  =\frac{T_1+...+T_C}{n}
\end{equation}

\begin{equation}\label{recall}
{\rm recall\;rate}  =(\frac{T_1}{n_1}+...+\frac{T_C}{n_C})/C
\end{equation}

where $T_c, c=1,...,C$ is the count of true samples in $c$th class, $n_c, i=1,...,C$ is the count of forecasting samples in $c$th class.

\begin{table*}[!ht]      
	\centering
	\caption{Description of data sets.}
	\label{Table2}
		\begin{tabular}[t]{c c c c c}
			\hline
			Datasets  &No. of Instances & No. of Features &No. of Classes & No. of Features after PCA\\
			\hline
			fac&2000&216&10&-\\
			fou&2000&76&10&-\\
			kar&2000&64&10&-\\
			pix&2000&240&10&-\\
			zer&2000&47&10&-\\
			mor&2000&6&10&-\\
			Yale&165&2000&15&100\\
			COIL20&1440&1024&20&200\\
			MNIST&2000&256&10&100\\
			USPS&1800&256&10&100\\
			\hline
		\end{tabular}
			\end{table*}

	\subsection{Parameters Setting}
	The performance of various feature extraction methods was evaluated by setting certain parameters in advance. First, the more appropriate default parameters for testing machine learning problems in Adam optimizer comprised $\alpha = 0.001$, $\beta_1 = 0.9$, $\beta_2 = 0.999$, and $\epsilon = 10^{−8}$. For all comparative algorithms, the search range of $k$ was set to $\{2,4,6,8,10\}$, whereas the range of $\sigma$ for CL-LPP, CL-LDA, and CL-LFDA was set as $\{0.01,0.1,1,10,100,1000\}$. In addition, the thermal parameter was calculated by $t=\|x_i-x^{(7)}_i\|\|x_j-x^{(7)}_j\|$, where $x^{(7)}_i$ and $x^{(7)}_j$ are the 7th nearest neighbors of $x_i$ and $x_j$, respectively.

					.

\section{Conclusion}
In this study, we proposed a unified feature extraction framework based on contrastive learning (CL-UFEF) that is suitable for both unsupervised and supervised learning. The proposed method defined the positive and negative pairs from a new perspective and subsequently applied them to the field of feature extraction. Concretelly, compared with the previous models based on contrastive learning, CL-UFEF does not need data enhancement, and it constructs positive and negative pairs based on two contrastive learning graphs (CLG), which will make the similar samples in the subspace more clustered. Compared with the traditional graph-based models, CL-UFEF is suitable for both unsupervised and supervised feature extraction, and it considers similar and dissimilar samples in unsupervised and supervised learning based on the CLG.\par


%

\ifCLASSOPTIONcaptionsoff
  \newpage
\fi



%
\bibliographystyle{IEEEtran}
\bibliography{bibfile}
\end{document}